\pdfoutput=1
\documentclass[11pt]{article}

\usepackage{acl}
\usepackage{times}
\usepackage{latexsym}
\usepackage{amsmath,array,graphicx}
\usepackage{kantlipsum}

\usepackage[T1]{fontenc}
\usepackage[utf8]{inputenc}

\usepackage{microtype}

\usepackage{todonotes}

\begin{document}

\title{Signal in Noise: Exploring Meaning Encoded in Random Character Sequences with Character-Aware Language Models}

% Author information can be set in various styles:
% For several authors from the same institution:
% \author{Author 1 \and ... \and Author n \\
%         Address line \\ ... \\ Address line}
% if the names do not fit well on one line use
%         Author 1 \\ {\bf Author 2} \\ ... \\ {\bf Author n} \\
% For authors from different institutions:
% \author{Author 1 \\ Address line \\  ... \\ Address line
%         \And  ... \And
%         Author n \\ Address line \\ ... \\ Address line}
% To start a seperate ``row'' of authors use \AND, as in
% \author{Author 1 \\ Address line \\  ... \\ Address line
%         \AND
%         Author 2 \\ Address line \\ ... \\ Address line \And
%         Author 3 \\ Address line \\ ... \\ Address line}

\author{Mark Bo Chu \\
  Columbia University  \\
%   Affiliation / Address line 2 \\
%   Affiliation / Address line 3 \\
  \texttt{mbc2165@columbia.edu} \\\And
  Bhargav Srinivasa Desikan \\
  \'{E}cole Polytechnique F\'{e}d\'{e}rale de Lausanne \\
%   Affiliation / Address line 3 \\
  \texttt{bhargav.srinivasadesikan@epfl.ch} \\ 
  \AND
  Ethan O.\ Nadler \\
  Carnegie Observatories  \\
  University of Southern California \\
%   Affiliation / Address line 2 \\
%   Affiliation / Address line 3 \\
  \texttt{enadler@carnegiescience.edu} \\\And
  D. Ruggiero Lo Sardo \\
   Sapienza University of Rome \\ 
   \texttt{losardor@gmail.com} \\ 
%   Affiliation / Address line 3 \\
  \AND
  Elise Darragh-Ford \\
  Stanford University \\
  %Kavli Institute of Astrophysics \\
  KIPAC \& Department of Physics \\
%   Affiliation / Address line 2 \\
%   Affiliation / Address line 3 \\
  \texttt{edarragh@stanford.edu} \\\And
  Douglas Guilbeault \\
  University of California, Berkeley \\
   Haas Business School \\ 
   \texttt{douglas.guilbeault@berkeley.edu} \\ 
%   Affiliation / Address line 3 \\
  }
 
\maketitle
%\vspace{-12mm}
\begin{abstract}

Natural language processing models learn word representations based on the distributional hypothesis, which asserts that word context (e.g., co-occurrence) correlates with meaning. We propose that $n$-grams composed of random character sequences, or \emph{garble}, provide a novel context for studying word meaning both within and beyond extant language. In particular, randomly generated character $n$-grams lack meaning but contain primitive information based on the distribution of characters they contain. By studying the embeddings of a large corpus of garble, extant language, and pseudowords using CharacterBERT, we identify an axis in the model's high-dimensional embedding space that separates these classes of $n$-grams. Furthermore, we show that this axis relates to structure within extant language, including word part-of-speech, morphology, and concept concreteness. Thus, in contrast to studies that are mainly limited to extant language, our work reveals that meaning and primitive information are intrinsically linked.

\end{abstract}

\section{Introduction}

What primitive information do character sequences contain? %According to \citet{deutsch2011beginning}, ``A writing system based on an alphabet can cover not only every word but every possible word in its language.'' 
Modern natural language processing is driven by the \textit{distributional hypothesis} \citep{firth1957synopsis}, which asserts that the context of a linguistic expression defines its meaning \citep{emerson2020goals}. 
Because existing words---which represent an extremely small fraction of the space of possible character sequences---appear in context together, the distributional paradigm at this level is limited in its ability to study the meaning of and information encoded by arbitrary character level $n$-grams (word forms). Furthermore, state-of-the-art computational language models operating within the distributional paradigm, such as BERT \citep{devlin2018bert}, are mainly trained on extant words.
Yet, a plethora of insights into language learning have emerged from inquiries into language beyond extant words, such as the grammatical errors and inference patterns that children exhibit when distinguishing extant words from non-linguistic auditory signals, including emotional expressions, auditory gestures, and other forms of paralinguistic speech \cite{yang2006infinite,carey2000origin}. We therefore propose that character $n$-grams (i.e., sequences of alphabetic characters) outside the space of extant language can provide new insights into the meaning of words and how they are represented by these models, beyond that captured by word and sub-word-based distributional semantics alone. We explore this by studying the embeddings of randomly generated character $n$-grams (referred to as \textit{garble}), which contain primitive communicative information but are devoid of meaning, using the CharacterBERT model \citep{el2020characterbert}. Such randomly generated character $n$-grams are textual analogues of paralinguistic vocalizations---vocal extra-speech sounds and noises.

Our analyses contribute to the growing understanding of BERTology \citep{rogers2020primer} by identifying a dimension, which we refer to as the \textit{information axis}, that separates extant and garble $n$-grams. This finding is supported by a Markov model that produces a probabilistic information measure for character $n$-grams based on their statistical properties. Strikingly, this information dimension correlates with properties of extant language; for example, parts of speech separate along the information axis, and word concreteness varies along a roughly orthogonal dimension in our projection of CharacterBERT embedding space. Although the information axis we identify separates extant and randomly generated $n$-grams very effectively, we demonstrate that these classes of $n$-grams mix into each other in detail, and that \emph{pseudowords}---i.e., phonologically coherent character $n$-grams without extant lexical meaning---lie between the two in our CharacterBERT embeddings.

This paper is organized as follows. We first discuss concepts from natural language processing, information theory, and linguistics relevant to our study. We then analyse CharacterBERT representations of extant and randomly generated character sequences and how the relation between the two informs the structure of extant language, including morphology, part-of-speech, and word concreteness. Finally, we ground our information axis in a predictive Markov language model.

\section{Modeling $n$-grams Beyond Extant Language}

Models in computational linguistics often represent words in a high-dimensional embedding space based on their co-occurrence patterns according to the distributional hypothesis \citep{landauer1997solution,mikolov2013distributed}. Embeddings that capture the semantic content of extant words are used for many natural language applications, including document or sentence classification \citep{kowsari2019text}, information retrieval and search \citep{mitra2018introduction}, language modelling and translation \citep{devlin2018bert}, language generation \citep{brown2020language}, and more \citep{jurafsky2009speech}. In these cases, vector operations performed on word %and character sequence
embeddings are used for higher-level tasks such as search or classification.

Word embeddings have largely concerned themselves with extant language---that is, commonly used words which carry consistent meaning---and thus cannot represent character $n$-grams outside of this space. The few models that encompass \emph{character} $n$-grams, which naturally include $n$-grams beyond extant words, often use RNNs \citep{mikolov2010recurrent} or encoder-decoder architectures \citep{sutskever2014sequence} to represent character-level sequences. 
In parallel, the ubiquitous use of Transformer models has led to studies of their inner representations, weights, and attention mechanism \citep{rogers2020primer,clark2019does}. Most Transformer models are trained using extant words and sub-words, largely focusing on their semantics and syntax; %, but are limited to extant words and extant language generation. 
however, %while most Transformer models use sub-word tokenization, 
some recent models operate at the character level, such as CharacterBERT \citep{el2020characterbert} and CharBERT \citep{ma2020charbert}. Strikingly, character-level models %noting that character inputs are sufficient 
excel at character-level tasks (e.g., spelling correction; \citealt{xie2016neural,chollampatt2018multilayer}) and perform comparably to word-level models at language-modelling tasks \citep{kim2016character}.

Character-level models are therefore an ideal tool for studying the information and meaning encoded in $n$-grams beyond the realm of extant language. Given that the current state-of-the-art is driven by Transformer-based models, throughout our study, we use the CharacterBERT model. CharacterBERT is uniquely suited for our study as it uses a CharacterCNN module \citep{peters2018deep} to produce single embeddings for any input token, built as a variant to BERT which relies on sub-word tokenization \citep{el2020characterbert}.

\section{Primitive Information and Meaning Beyond Extant Language}

Before presenting our results, we discuss general characteristics of the space beyond extant words; we reiterate that this space is missed by word and sub-word-based models. Due to CharacterBERT's use of English characters, we restrict our analysis to English character $n$-grams, and we study the properties of CharacterBERT embeddings including English-based $n$-grams outside of extant language. By studying CharacterBERT's representations of meaning encoded in $n$-grams that do not appear in consistent (or any) context in its training data, our framework goes beyond the traditional distributional hypothesis paradigm. In this way, we seek to understand core properties of information encoded in $n$-grams beyond their lexicalized semantics by simultaneously studying $n$-grams that contain different types of information.\footnote{In analogy, the theory of ensemble perception in developmental psychology offers a framework to understand the human ability to understand the `gist' of multiple objects at once \citep{sweeny2015ensemble}.}%In our study, through CharacterBERT,

We use randomly generated character sequences to create  $n$-grams that contain primitive information but no meaning. We adapt Marr's notion of primitive visual information for primitive textual information
%descriptions for a class of visual computational information, where lower-level visual features inform holistic visual representations, like in the case of edge detection informing object representation 
\citep{marr1980theory}, and make the analogue between vision and language because information is substrate independent \citep{deutsch2015constructor}. In our case, primitive textual information is lower-level communicative information which is present in both text with and without meaning. Being textual, our randomly generated $n$-grams are not bound by the constraints of human speech, and may be phonologically impossible; these garble $n$-grams may be seen as an example of textual noise.

%At token level, there is infinite variety; \textbf{we limit our sample of random character $n$-grams by copying the character-length distribution of the 40,000 most commonly-used English words.}

In the following subsections, we provide three examples of language---distorted speech, paralanguage, and pseudowords---which motivate our study of character-level embeddings for randomly generated character $n$-grams. We then describe the complementary information encoded by word morphology.%, which we call \textit{garble}.

\subsection{Distorted Speech}

In popular use, ``garble'' refers to a message that has been distorted (garbled), such as speech where meaning is corrupted by phonological distortions. For example, the phrase “reading lamp” may become “eeling am” when garbled. Garbled speech contains lesser, or zero, meaning compared to ungarbled speech, but the signal of speech media is nonetheless present as information, which according to \citet{shannon1951redundancy} may contain no meaning at all. Garbled speech satisfies the classical five-part definition of communication provided by \citet{shannon2001mathematical}; an \textit{information source} (speaker) can \textit{transmit} (verbalize) an informationally primitive message through the \textit{channel} of speech media through the \textit{receiver} (ears) to the \textit{destination} (listener).

\subsection{Paralanguage}

 Paralinguistic vocalizations are specifically identifiable sounds beyond the general characteristics of speech \citep{noth1990handbook} and present another example of communication beyond lexicalized semantics. Paralinguistic vocalizations include \textit{characterizers}, like moaning; and \textit{segregates}, like ``uh-huh'' for affirmation. The border between such paralinguistic vocalizations and lexicalized interjections with defined meanings is ``fuzzy'' \citep{noth1990handbook}.%\footnote{Note, semiotician Winifried Noth does not define his use of "fuzzy" as mathematical or metaphoric}. 

 %Note that distinguishing between paralinguistic vocalizations and garbled speech may be unobservable. E.g. A person choking may wish to verbalize a call for help using lexical speech but are impeded from articulating the phonetics of lexicalized vocabulary. When garbled, the lexical speech, “Help me!” may become the paralinguistic vocalization, “Eh eh!”  communicating primitive speech information without semantics. 

\subsection{Pseudowords}

Pseudowords are phonologically possible character $n$-grams without extant lexical meaning. Wordlikeness judgments reveal that human distinctions between pseudowords and phonologically impossible nonwords are gradational
%, with  subjective morphophonology described as an experience-based generalization of probabilities in the lexicon 
\citep{needle2020phonological}. As a unique informational class, pseudowords have been used in language neuronal activation studies \citep{price1996demonstrating}, infant lexical-semantic processing \citep{10.1162/089892905774589172}, in poetry through nonsense \citep{ede1975nonsense}, and in literary analyses \citep{lecercle2012philosophy}. Pseudowords can also elicit similar interpretations and associations across independent participants \citep{davis2019horgous}.

To consider pseudowords generatively, it is helpful to note that an alphabetic writing system covers not only every word but every possible word in its language \citep{deutsch2011beginning}; pseudowords can thus be thought of as possible-but-uninstantiated (counterfactual) extant words---e.g., ``cyberspace'' was a pseudoword before the internet. We embed randomly generated pseudowords into our model to study their information content and relation to both extant words and randomly generated $n$-grams.

\subsection{Morphology}

Morphology deals with the systems of natural language that create words and word forms from smaller units \citep{10.7551/mitpress/4775.003.0005}. Embedding spaces and the distributional hypothesis offer insights into the relationship between character combination, morphology and semantics. Notably, morphological irregularities complicate the statistics of global character-level findings in the embedding space, like through \emph{suppletion}---where word forms change idiosynchratically e.g. \textit{go}'s past tense is \textit{went}, or \emph{epenthesis}---where characters are inserted under certain phonological conditions e.g. fox pluralizes as fox\textit{e}s \citep{10.7551/mitpress/4775.003.0005}; so too do the multiple `correct' spellings of pseudowords under conventional phoneme-to-grapheme mappings \citep{needle2020phonological}. 
Distinctions between morphological phenomena can also be hard to define; for example, the boundary between derivation and compounding is ``fuzzy'' \citep{10.7551/mitpress/4775.003.0005}.%\footnote{Note again, computational linguist Harald Trost does not define his use of "fuzzy" as mathematical or metaphoric}.

\begin{figure*}[t!]
    \centering
    \includegraphics[width=0.9\textwidth,trim={0 0in 0 4in}]{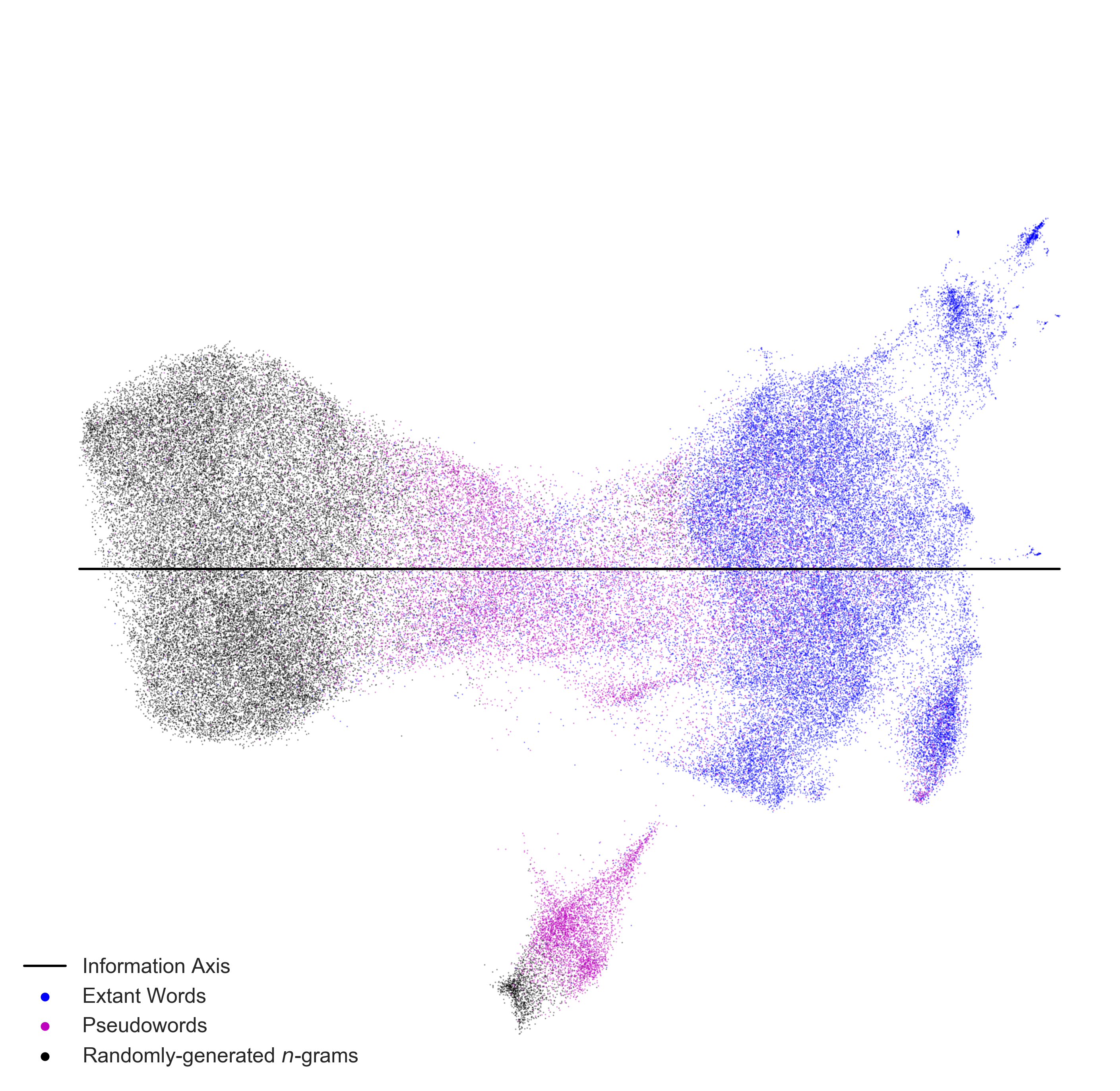}
    \caption{UMAP projection of CharacterBERT embeddings for extant words (blue), pseudowords (magenta), and randomly generated character $n$-grams (black). The solid black line shows the information axis that we define in this work. The bottom-most cluster of random and pseudoword character $n$-grams is comprised of character $n$-grams ending in ``s'', and the top-most clusters of extant words are comprised of compound words.}
    \label{fig:fig1}
\end{figure*}

\section{Character-Level Language Models for Information Analysis}

%We see there are many ways to encode information in the linguistic sense, even outside of extant words. 
As described above, state-of-the-art language models serve as a tool to study meaning as it emerges though the distributional hypothesis paradigm. Existing work on the analysis of Transformers and BERT-based models have explored themes we are interested in, such as semantics \citep{ethayarajh2019contextual}, syntax~\citep{goldberg2019assessing}, morphology \citep{hofmann2020dagobert, hofmann2021superbizarre}, and the structure of language \citep{jawahar2019does}. However, all of this work limits itself to the focus of extant words due to the word and sub-word-based nature of these~models.

We study the structure of the largely unexplored character $n$-gram space which includes extant language, pseudowords and garble character $n$-grams, seen through the representations created by CharacterBERT, as follows. To explore how the character $n$-gram space is structured in the context of character based distributional semantics, we embed 40,000 extant English words, 40,000 randomly generated character $n$-grams, and 20,000 pseudowords. We choose the 40,000 most used English words that have been annotated for concreteness/abstractness ratings \citep{brysbaert2014concreteness}. Randomly generated character $n$-grams are forced to have a string length distribution that matches the corpus of extant words we analyze. To generate pseudowords, we use a popular pseudoword generator.\footnote{\href{http://soybomb.com/tricks/words/}{http://soybomb.com/tricks/words/}}

The CharacterBERT \citep{el2020characterbert} general model has been trained on nearly 40 GB of Reddit data using character sequences. We leverage this model to create representations of character $n$-grams that may not have been seen in the training data. This allows us to use the resulting 512 dimensional embeddings for exploration via visualisation, topology modelling via distances and projections, and classification error analysis.

\subsection{Identifying the Information Axis}

To guide our exploration of the high-dimensional topology of the resulting embeddings, we use the UMAP dimensionality reduction technique \citep{mcinnes2018umap}. UMAP creates a low-dimensional embedding by searching for a low-dimensional projection of the data that has the closest possible equivalent fuzzy topological structure as the original representations, thereby preserving both local and global structure. In Appendix \ref{appendixa}, we demonstrate that our key results are not sensitive to this choice of dimensionality reduction method.

We use the UMAP embeddings to extract an \textit{information axis} that captures most variance among extant and randomly generated $n$-grams. To assign $n$-grams an `information axis score,' we minmax-normalize the UMAP coordinates along this axis. Thus, our information axis establishes a link between extant language and garble, thereby connecting meaning and primitive information. Figure \ref{fig:fig1} shows how CharacterBERT embeddings of extant, pseudoword, and randomly generated character $n$-grams arrange themselves in this space.

\subsection{Statistical Properties of $n$-grams Along the Information Axis}

%(\textbf{Ethan to double check/edit this section})

We perform several statistical tests to differentiate between categories of character $n$-grams along the information axis. First, Table \ref{tab:table1} lists the median and standard deviation of minmax-normalized position along the information axis, demonstrating that extant words, pseudowords, and garble are clearly separated. Note that the scatter within each $n$-gram class is much smaller than the distances between classes, indicating that our results are robust to variations in the garble and pseudoword samples.

\begin{figure}[t]
    % \centering
    \hspace{-3mm}
    \includegraphics[width=0.5\textwidth]{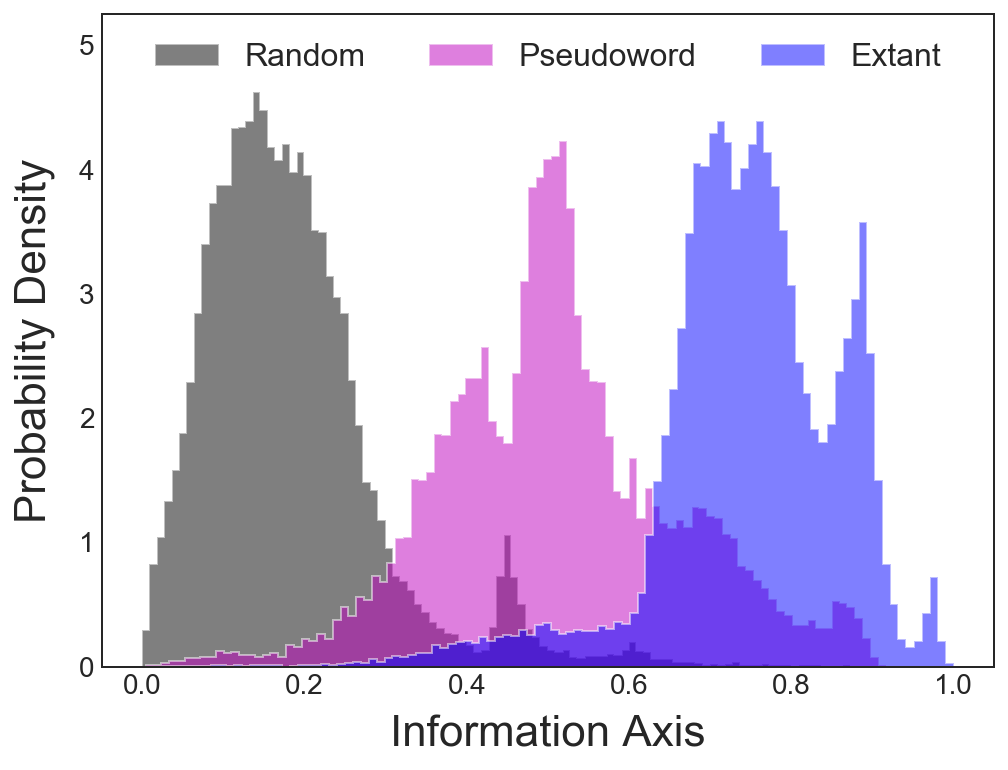}
    \caption{Probability density of CharacterBERT embeddings for extant words (blue), pseudowords (magenta), and randomly generated character $n$-grams (black) as a function of minmax-normalized position along the information axis shown in Figure \ref{fig:fig1}.}
    \label{fig:fig2}
\end{figure}

Next, we use the Kolmogorov-Smirnov (KS; \citealt{massey1951kolmogorov}) two-sample test to assess differences between the information axis distributions of our $n$-gram classes. All of the KS tests very significantly indicate differences between types of character $n$-gram and parts of speech along the information axis ($p\ll 0.001$). Furthermore, the KS statistic score is 0.94 for (extant, random), 0.83 for (pseudoword, random), and 0.70 for (extant, pseudoword), indicating that extant and random $n$-grams differ most significantly along the information axis (consistent with Figures \ref{fig:fig1}--\ref{fig:fig2}).

\begin{table}[t!]
\resizebox{\columnwidth}{!}{%
\begin{tabular}{c*{2}{>{}c<{}}}
  Character $n$-gram type  &  Information Axis position     \\
    \hline
Extant  & $0.75\pm 0.12$ \\
Noun  & $0.74\pm 0.12$ \\
Verb  & $0.72\pm 0.09$  \\
Adjective  & $0.76 \pm 0.11$  \\
Adverb  & $0.87\pm 0.09$  \\
Pseudoword & $0.50\pm 0.15$ \\      
Random & $0.17\pm 0.11$
\end{tabular}
}
\caption{Median and standard deviation of minmax-normalized position along the information axis shown in Figure \ref{fig:fig1}, for extant words (including parts of speech), pseudowords, and randomly generated $n$-grams.\label{tab:table1}}
\end{table}

\subsection{Hyperplane Classifier} 

The visualisation of the character $n$-grams suggests that a hyperplane classifier is suitable for separating extant words %pseudowords, 
and garble. We use a support vector machine \citep{cortes1995support} trained on half of our 40,000 commonly-used extant words and half of our computer-generated garble to classify unseen extant, garble and pseudoword character $n$-grams. We use this method to explore the information axis in the high-dimensional embedding space.

The classifier achieves an accuracy of 98.9\% on unseen extant language and garble character $n$-grams, suggesting we can learn about the  embeddings through error analysis. 

In particular, we found similarities among extant words classified as garble. $74.4\%$ (270/363) were compound or derivative words, similar to many extant language terms that lie near the midpoint of the information axis. $19\%$ (69/363) were foreign words like ``hibachi'' or dialect words like ``doohickey.'' 

The garble classification errors---garble classified as extant language---were in small part due to our randomization method inadvertently creating extant language labelled as garble, accounting for $9.5\%$ (36/377) errors we identify. The garble classified as extant language mostly contained phonologically impossible elements, though some were pseudowords.

When pseudowords were forcibly classified into extant or garble character $n$-grams, more pseudowords were classified as extant language than garble (12894 as extant to 7106 as garble). Labelling affirms these intuitions, with pseudowords like ``flought'' looking intuitively familiar and being readable. Given CharacterBERT's massive Reddit training data, typos and localized language may account for the classifier's tendency to classify pseudowords as extant language. Also, our embedding space only uses the 40,000 most common English words out of 208,000 distinct lexicalized lemma words \citep{brysbaert2016many}, which may impact spatial structure if included.

\begin{figure*}[t]
    % \centering
    \includegraphics[width=0.5\textwidth,trim={0 0 0 4in}]{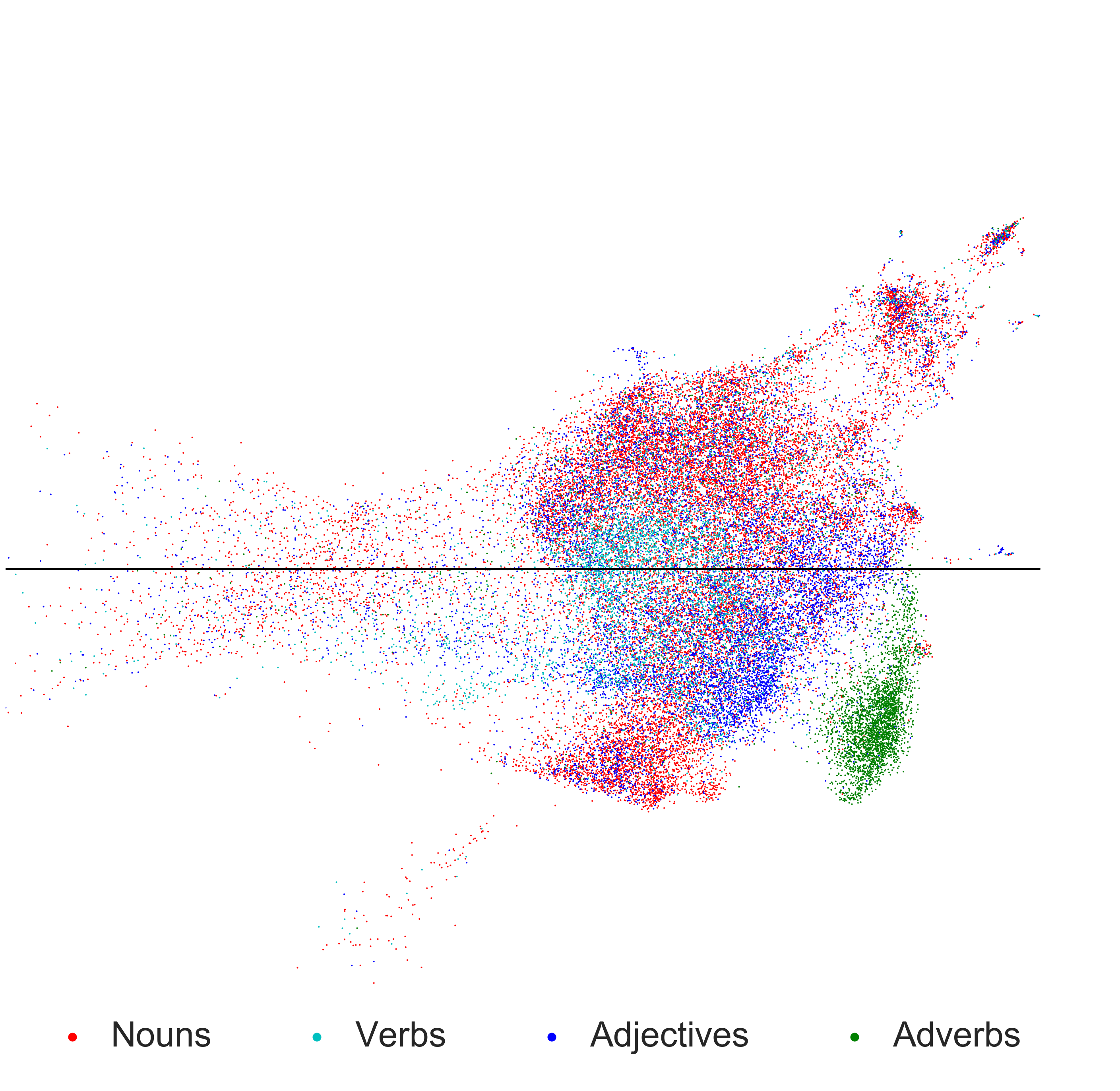}
    \includegraphics[width=0.5\textwidth]{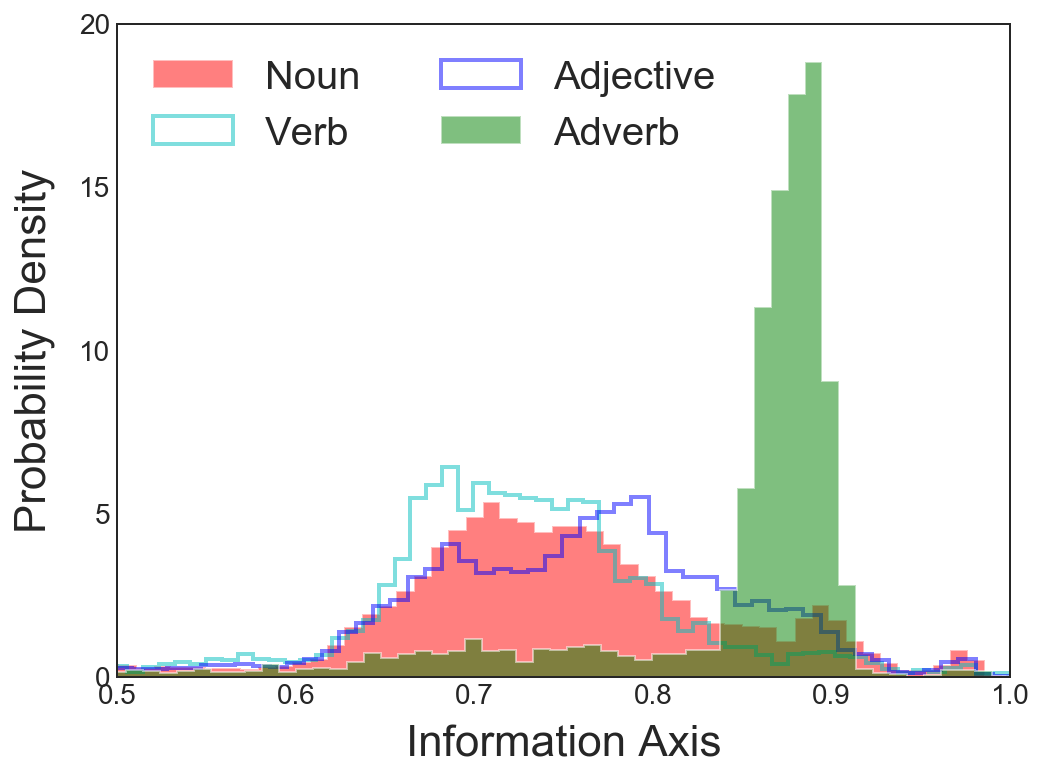}
    \caption{\emph{Left panel}: UMAP projection of CharacterBERT embeddings for extant words split by part-of-speech into nouns (red), verbs (cyan), adjectives (blue), and adverbs (green). \emph{Right panel}: Probability density of extant words, split by part-of-speech, as a function of minmax-normalized position along the information axis shown in Figure \ref{fig:fig1}.}
    \label{fig:fig3}
\end{figure*}

%%%%%%%%%%%%%%%%%%%%%%%%%%%%%%%%%%%%%%%

\section{Structure of Extant Words along the Information Axis}

We use this section to discuss the structure of language across the information axis derived from our low-dimensional UMAP space. We structure our analysis across this axis as it organises the relative structure of extant words vs. randomly generated character $n$-grams, while also distinguishing internal structure within the extant word space.

\begin{figure*}[t]
    % \centering
    \vspace*{1.5cm}
    \includegraphics[width=0.5\textwidth,trim={0 0 0 4in}]{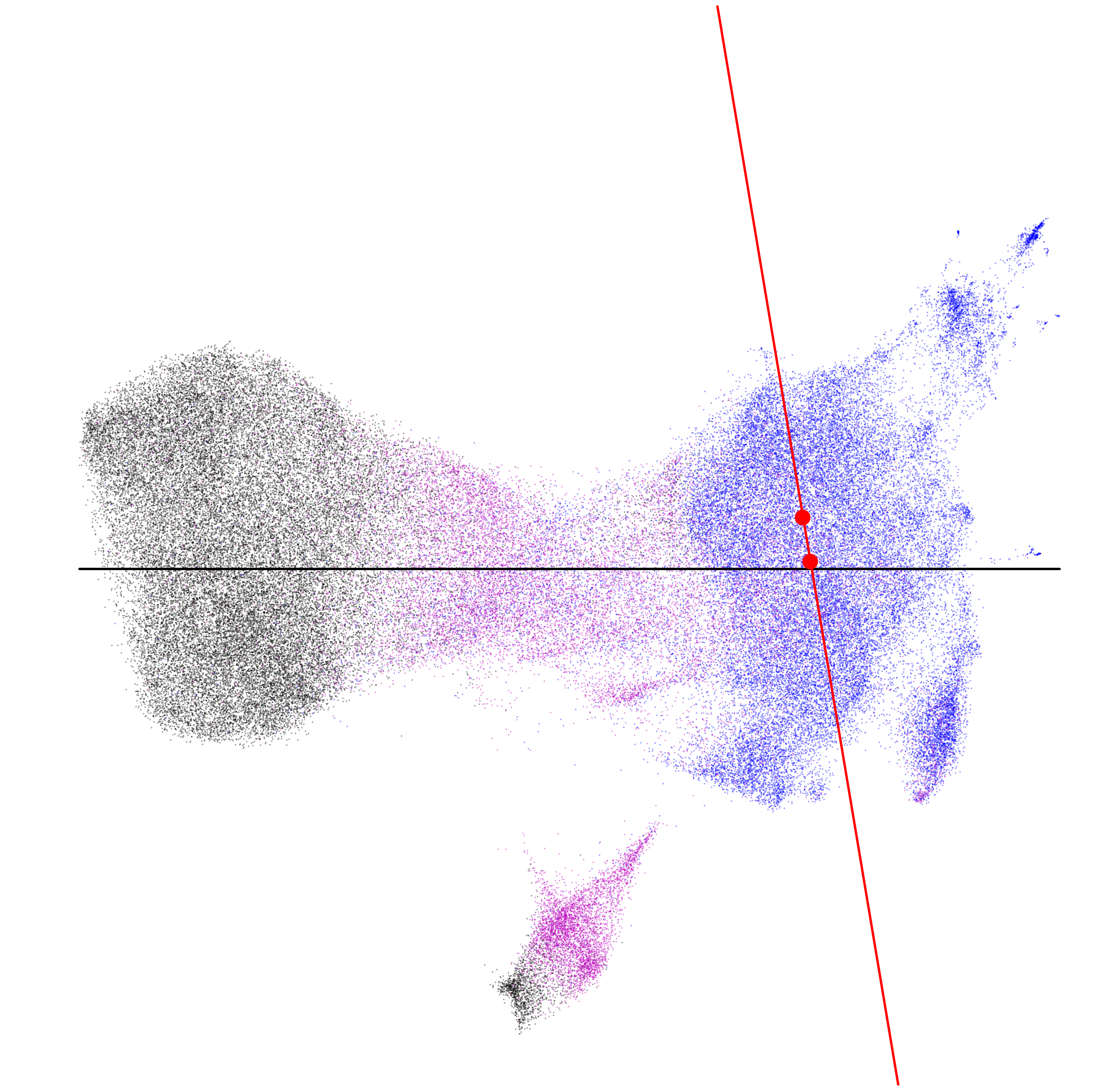}
    \includegraphics[width=0.5\textwidth,trim={0 0 0 4in}]{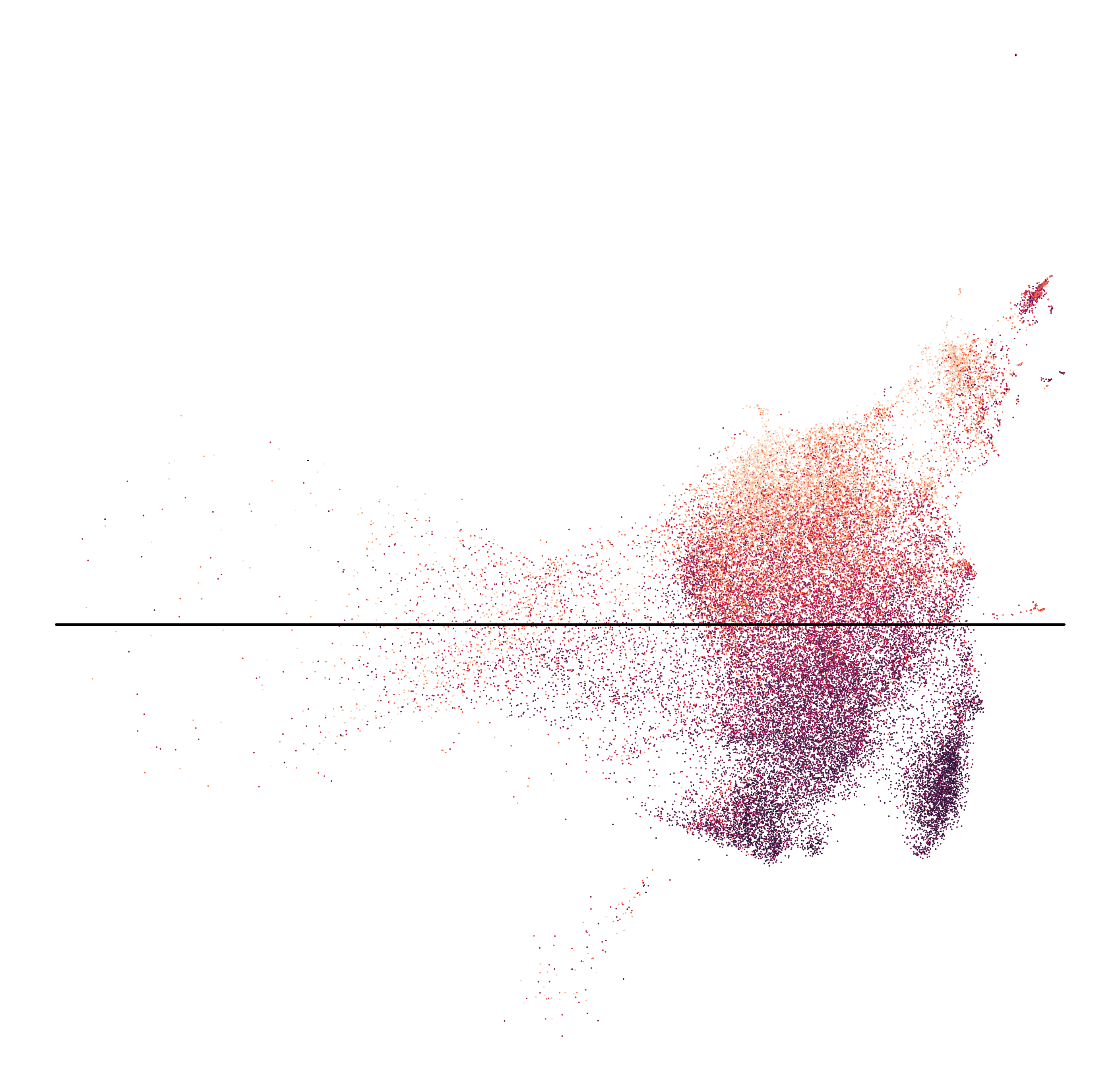}
    \caption{\emph{Left panel}: UMAP projection of CharacterBERT embeddings for extant words (blue), pseudowords (magenta), and randomly generated character $n$-grams (black). The solid black line shows the information axis that we define in this work, and the red line shows the axis that captures variability in word concreteness, computed by connecting the unweighted average UMAP position for extant words with that weighted by minmax-normalized concreteness (red dots). \emph{Right panel}: UMAP of only extant words, colored by minmax-normalized concreteness, with lighter colors indicating more concrete words.}
    \label{fig:fig4}
\end{figure*}

\subsection{Extant vs.\ Pseudowords vs.\ Garble}

At the scale of global structure, the information axis highlights that extant words are separated from randomly generated character $n$-grams (Figure \ref{fig:fig1}). We note that the midpoint of all character $n$-gram classes is 0.5 on our information axis. Pseudowords populate the region near the midpoint of the information axis, and also overlap with both extant English and garble character $n$-grams (Figure \ref{fig:fig2}). There is no distinct boundary between the three classes of $n$-grams, consistent with both morphological descriptions of compound and derivational words and descriptions of paralanguage as ``fuzzy.'' This global structure---and the structure internal to extant language (Figure \ref{fig:fig3})---goes beyond the distributional hypothesis by including $n$-grams that do not appear in consistent (or any) contexts, like pseudowords and garble. Pseudowords lie between extant and garble character $n$-grams, but there is no distinct boundary between pseudowords and the other classes of $n$-grams.%We consider pseudowords to also be  the context for extant language.

Extant language, pseudoword, and garble regions have different internal structure (Figure \ref{fig:fig1}). The garble region has comparatively less structure than the extant language region, though there is some internal variation, notably a cluster of character $n$-grams ending in the character ``s'' separated from the main garble region. We qualitatively explore the classes of garble and pseudoword embeddings revealed by our analysis in Appendix \ref{appendixb}, which includes supplementary discussion of the potential relevance of these findings for linguistic theory.

% However, the garble region's relative position to the extant language region informs the varied structures within the extant language region via the information axis. 

\subsection{Parts of Speech and Morphology}

% When grouped by parts of speech (Fig.\ref{fig:fig3}), adverbs lie farthest from the midpoint of the information axis. Nouns, adjectives, and verbs lie roughly equidistant from the midnight of the information axis.

 In our UMAP projection, detailed structure emerges for extant words split by part-of-speech (Figure \ref{fig:fig3}). In particular KS statistics between all part-of-speech pairs significantly indicate that their distributions differ along the information axis. Furthermore, KS statistic values are 0.12 for (noun, verb), 0.11 for (noun, adjective), 0.64 for (noun, adverb), 0.22 for (verb, adjective), 0.72 for (verb, adverb), and 0.64 for (adjective, adverb). This suggests that adverbs are most cleanly separated from other parts of speech along the information axis (consistent with Figure \ref{fig:fig3}), which may indicate that morphemes such as affixes have important effects in embedding space. A detailed investigation is beyond the scope of this paper and may require analyses through alternative heuristics such as pseudomorphology and lexical neighborhood density \citep{needle2020phonological}. 

Many extant words near the midpoint of the information axis are, or may be, compound words; the boundary between derivative and compound words is thought to be fuzzy because many derivational suffixes developed from words are frequently used in compounding \citep{10.7551/mitpress/4775.003.0005}. Both derivative and compound words populate other spaces of the extant language region, but conflicting definitions hamper straightforward statistical analysis. 

Morphological traits such as adjectival suffixes $-ness$, $-ism$, and $-able$, or the adverbial suffix $-ly$ correlate to clear embedding mappings, but the boundaries for morphological classes are not distinct. Garble ending in ``s'' occupies a closer region to extant language than most other garble, arguably due to the semantic associations of ending in ``s'' (e.g. regarding pluralization) derived from the suffix $-s$. Note, morphological heuristics like affixation apply to lexicalized words but not garble. Pseudowords ending in ``s'' share that region of garble ending in ``s'', however, such seemingly plural pseudowords tend closer to extant language, reflecting the notion that word form similarity increases with semantic similarity \citep{dautriche2017wordform}. Given the fuzziness of morphology and the opaqueness of English spelling \citep{needle2020phonological}, pseudowords ending in ``s'' may or may not be due to affixation. 

\begin{figure*}[h]
    % \centering
    \includegraphics[width=0.5\textwidth]{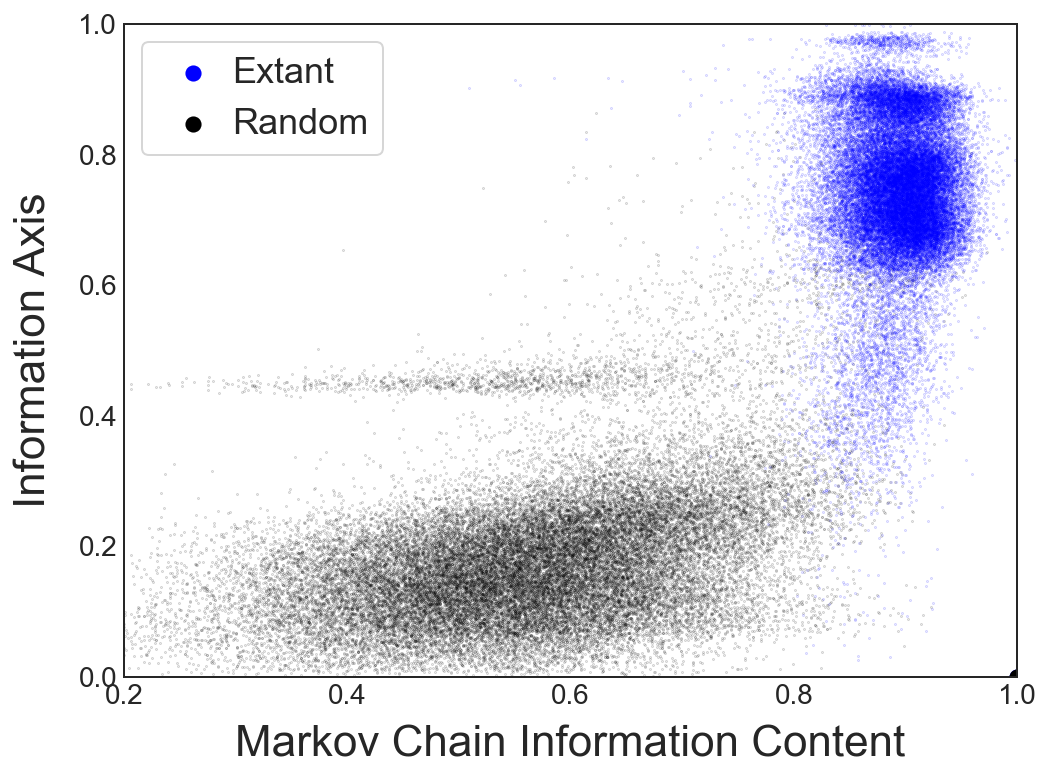}
    \includegraphics[width=0.5\textwidth]{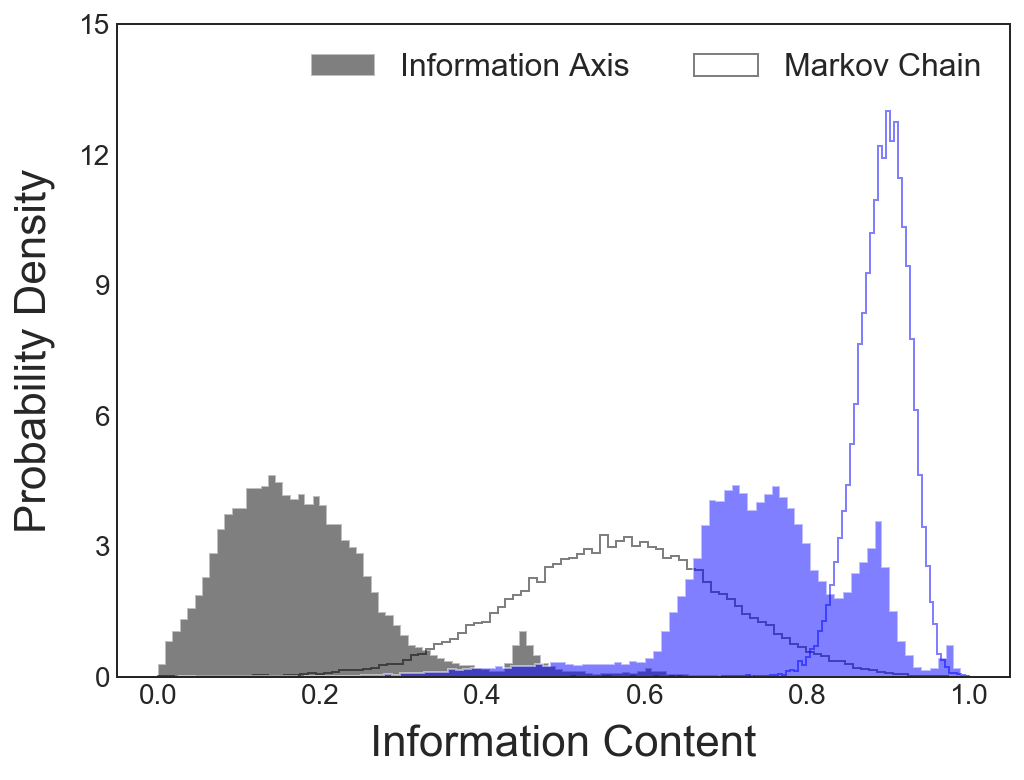}
    \caption{\emph{Left panel}: Minmax-normalized position along the information axis shown in Figure \ref{fig:fig1} vs.\ minmax-normalized information content from our Markov Chain model, for extant words (blue) and randomly generated character $n$-grams (black). \emph{Right panel}: Probability density of minmax-normalized information content measures from our UMAP projection (filled histograms) and Markov Chain model (unfilled histograms).}
    \label{fig:fig5}
\end{figure*}

%\textbf{(Mark can you expand on this section?)}

\subsection{ Concreteness/Abstractness}

The internal positioning of different parts-of-speech within the extant language space of our low-dimensional UMAP projection suggests that the representations also capture notions of concreteness (e.g nouns) and abstractness (e.g adverbs) which we explore by projecting concreteness scores from the \citep{brysbaert2014concreteness} study. We calculate the center of extant UMAP coordinates with no weighting and with weighting by minmax-normalized concreteness and used those points to define a \textit{concreteness axis}, which demonstrates that concreteness varies in a direction roughly orthogonal to our information axis (see Figure \ref{fig:fig4}). The bootstrap-resampled angle distribution between information and concreteness axes is $86.6 \pm 1.2$ degrees.

Thus, the information axis and word concreteness capture two crucial and largely distinct aspects of the many latent features underlying CharacterBERT representations. This finding is particularly relevant in light of recent work showing not only that word concreteness is a psychologically rich dimension that shapes semantic processing \citep{brysbaert2016many,guilbeault2020color}, but also that word concreteness is surprisingly effective at enriching the predictive capacities of word embedding models, such as for the purpose of automated metaphor detection \cite{desikan2020comp}. We leave a detailed investigation of this finding, including its relation to the visual information \citep{brysbaert2016many} carried by concrete and abstract words, to future work.

\subsection{Markov Chain Model}

We also create a language model using the Prediction by Partial Matching (\textsc{ppm}) variable order Markov model (\textsc{vomm}) to estimate the probability of each of these character $n$-grams \citep{begleiter2004prediction}. The model calculates the \emph{logpdf} for each character $n$-gram in which more commonly occurring character $n$-grams have a lower score, and less commonly occurring character $n$-grams receive a higher score. The model is trained on extant words, then used to score all of the extant, pseudowords and garble character $n$-grams. We use this score to capture the likelihood of character $n$-grams in our character sequence space (Figure \ref{fig:fig5}).

These Markov model values correlate with our information axis measure. In particular, the Spearman correlation coefficient between information axis and Markov chain information content is 0.4 (highly significant) for randomly generated $n$-grams, and 0.007 (not significant) for extant words. Thus, for random character $n$-grams, our information axis measure is correlated with statistical properties of the character $n$-grams from the Markov model (see the left panel of Figure \ref{fig:fig5}). However, our information axis measure more clearly separates extant and garble $n$-grams, indicating that it incorporates information beyond purely statistical properties of $n$-gram classes (see the right panel of Figure \ref{fig:fig5}). This suggests that the CharacterBERT model learns information beyond character-level statistical information, even for $n$-grams that never explicitly appear in the training data.

\section{Discussion and Conclusion}

Using the CharacterBERT model, we embedded a large corpus of character level $n$-grams outside of extant language to study how the primitive information they contain relates to the semantic information carried by extant language. %We did so based on the distributional hypothesis, which asserts that the context of words defines their meaning.
The key findings of this paper are:
\begin{enumerate}
    \item Extant words and randomly generated character $n$-grams are separated along a particular axis in our UMAP projection of CharacterBERT embedding space (Figures \ref{fig:fig1}--\ref{fig:fig2}); %structure in our novel character $n$-gram space which we populate with extant language, pseudoword and randomly generated garble character $n$-grams (Fig.\ref{fig:fig1}).
    %\item This structure is best highlighted by the information axis, showing extant language as separate from randomly generated garble character $n$-grams. We center the midpoint of the information axis at 0.5 (Fig.\ref{fig:fig2}).
    \item Pseudowords lie between extant and randomly generated $n$-grams along this axis, but there is no distinct boundary between these classes of $n$-grams (Figures \ref{fig:fig1}--\ref{fig:fig2});
    %and the approximate boundary of pseudowords corroborates the use of "fuzzy" as a descriptor in morphology and linguistics (Fig.\ref{fig:fig2}).
    \item The structure of CharacterBERT embeddings of extant language, including structure based on part-of-speech and morphology, is correlated with the information axis (Figure \ref{fig:fig3});
    %The information axis highlights the internal structure of extant language. Speech classes form distinct regions, with adverbs located farthest from garble (Fig.\ref{fig:fig3}).
    \item Word concreteness varies along a dimension that is roughly orthogonal to the information axis in our UMAP projection (Figure \ref{fig:fig4});
    \item Separation between extant and randomly generated $n$-grams captured by CharacterBERT is correlated with and more coherent than that based purely on the statistical properties of $n$-grams (Figure \ref{fig:fig5}).
    %We use concreteness/abstractness scores of extant language to project a Concreteness Axis, which lies roughly orthogonal to the information axis (Fig.\ref{fig:fig4}).
\end{enumerate}

These findings suggest that character-based Transformer models are largely able to explore the relation between extant words and randomly generated character strings. %, and thus serve as a way to explore the bridge between these two classes of $n$-grams. 
In particular, character-level models capture complex structure in the space of words, pseudowords, and randomly generated $n$-grams. 
These findings are consistent with work suggesting that character-level and morpheme-aware representations are rich in meaning, even compared to  word or sub-word models \citep{al2019character,el2020characterbert,ma2020charbert, hofmann2020dagobert, hofmann2021superbizarre}.
%we learn that there is a high level of complexity in the representations of character-level models, supporting work that 
% suggesting that character-level and morpheme-aware models \citep{hofmann2020dagobert, hofmann2021superbizarre} are as powerful (if not more) than word or sub-word models \citep{al2019character,el2020characterbert,ma2020charbert}. 

%\subsection{Limitations and Future Work}

Our study is limited to extant words in English and randomly generated character $n$-grams using the English alphabet. Given the unique impact of a specific language and alphabet on representation spaces, there is motivation to see whether the relationships we identify generalise to other languages and alphabets. Finally, we reiterate that our analysis was limited to the last embedding layer of the CharacterBERT model; future work may focus on weights in earlier layers, including attention mechanisms explored by other BERTology studies  \citep{clark2019does,jawahar2019does}. By only analysing the final embedding layer, we study the `psychology' of such character-level models; in analogy, much may be gained by studying the `neuroscience' of such models encoded in their attention weights \citep{wang2020curious}.

Our study also has important practical implications for the widespread use of pseudowords as an experimental tool in psycholinguistic research. Pseudowords are frequently used as stimuli to observe the psychological and neurocognitive processes underlying the interpretation of novel words \cite{price1996demonstrating, stark2000repetition, keuleers2010wuggy, lupyan2015meaningless, davis2019does}. However, the lion’s share of this research treats all pseudoword stimuli as equivalent in their novelty, based on \textit{prima facie} human judgments. By contrast, our method shows that not all pseudowords are created equal. Due to various features of character sequences, including morphological structure, some pseudowords encode disproportionately more information according to character-aware language models, and are therefore represented as significantly more similar to extant words, whereas other pseudowords are recognized by these models as random character sequences. This variation is especially striking given that the algorithms used to generate pseudowords are highly constrained and designed to produce morphologically coherent words \cite{keuleers2010wuggy}; that some pseudowords are evaluated as random by CharacterBERT reveals not only asymmetries in the coherence of pseudowords that may be of psychological relevance, but also assumptions and limitations in terms of which morphological units CharacterBERT and related models recognize as signatures of extant words. Our study thus provides a quantitative method for evaluating pseudoword plausibility, without relying on variable human judgments, while also revealing insights into key differences between how humans and contemporary language models evaluate the plausibility of pseudowords.

To allow for further explorations and replicability, we release all of our data and code on GitHub\footnote{https://github.com/comp-syn/garble}. Our findings reveal new avenues for future work using character-aware embeddings of extant, pseudoword, and garble $n$-grams, including analyses of nonsense poetry like Lewis Carroll's ``Jabberwocky'' or of the innovative idiosyncrasies of rap lyricists and graffiti artists. The embeddings we study may also complement philological studies (especially if dynamic analyses are employed), as well as research into novel category formation \cite{lupyan2015meaningless,guilbeault2021experimental}. Also, language acquisition studies of the distinction between language and noise may benefit from character-level embeddings beyond the realm of extant language \cite{yang2006infinite,carey2000origin}. By investigating a broadened embedding space to include randomly generated $n$-grams, we found new structures of meaning through the context of meaningless information; further studies may extend our garble-based approach across different media and modes to contribute to more general understandings of human meaning.

\bibliography{anthology,custom}
\bibliographystyle{acl_natbib}

\clearpage

\appendix

\section{Robustness to Alternative Dimensionality Reduction Techniques}
\label{appendixa}

\begin{figure*}[t!]
    \centering
    \includegraphics[width=0.9\textwidth,trim={0 0in 0 2.5in}]{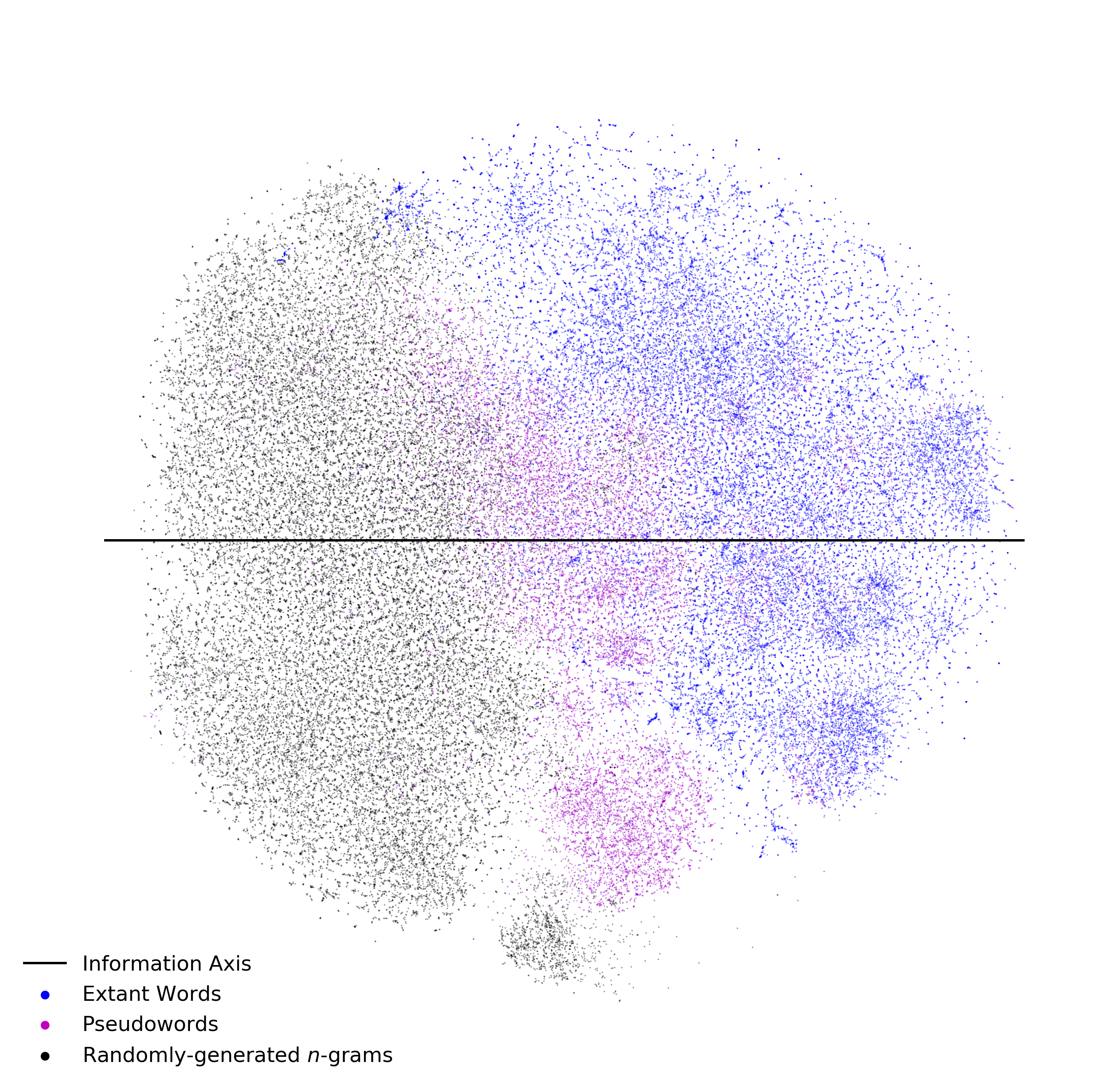}
    \caption{t-SNE projection of CharacterBERT embeddings for extant words (blue), pseudowords (magenta), and randomly generated character $n$-grams (black). The solid black line shows the information axis that we define in this work. The bottom-most cluster of random and pseudoword character $n$-grams is comprised of character $n$-grams ending in ``s.''}
    \label{fig:fig6}
\end{figure*}

Our main analyses use the UMAP algorithm to project garble, pseudoword, and extant word CharacterBERT embeddings into an interpretable, low-dimensional space. Here, we demonstrate that our key results are not sensitive to this choice of dimensionality reduction technique by recreating our findings using t-SNE, a popular alternative to UMAP. Figure \ref{fig:fig6} shows the extant, pseudoword, and garble embeddings resulting from the \texttt{scikit-learn} t-SNE algorithm (run with $n_{\mathrm{components}}=2$ and $\mathrm{perplexity}=10$). The qualitative structure is unchanged relative to the UMAP embedding shown in Figure \ref{fig:fig1}: garble and extant $n$-grams are separated along a new information axis that captures roughly the same amount of variance as our original UMAP information axis, and pseudowords embeddings connect these two clusters. Furthermore, some particular aspects of the UMAP structure are preserved, including a distinct cluster of garble and pseudoword $n$-grams ending in ``s'' near the bottom of Figure \ref{fig:fig6}. In general, the separation among t-SNE $n$-gram clusters is somewhat less distinct compared to the UMAP case, which we attribute to UMAP's better preservation of global structure \cite{mcinnes2018umap}.

The results of this t-SNE projection are also quantitatively consistent with our main findings. In particular, the UMAP information axis summary statistics presented in Table \ref{tab:table1} become $0.70\pm 0.15$, $0.54\pm 0.13$, and $0.26\pm 0.12$ for extant, pseudoword, and randomly generated $n$-grams, respectively; these results are all consistent with our UMAP results at the $1\sigma$ level. Similarly, KS two-sample tests between the extant, pseudoword, and garble information axis distributions all remain highly significant ($p\ll 0.001$), and their ordering is consistent with our UMAP results: the t-SNE information axis KS statistic scores are $0.86$ for (extant, random), $0.76$ for (pseudoword, random), and $0.50$ for (extant, pseudoword). Relative to our fiducial UMAP results, the slightly larger scatter for the information axis summary statistics and the slightly weaker KS statistic scores are consistent with the increases in scatter orthogonal to the information axis in the t-SNE projection (Figure \ref{fig:fig6}) relative to the UMAP projection (Figure \ref{fig:fig1}). Thus, our main results are not sensitive to the dimensionality reduction method employed.

\section{Global Structure of Garble and Pseudoword Embeddings}
\label{appendixb}

\begin{figure*}[t!]
    \centering
    \includegraphics[width=0.9\textwidth]{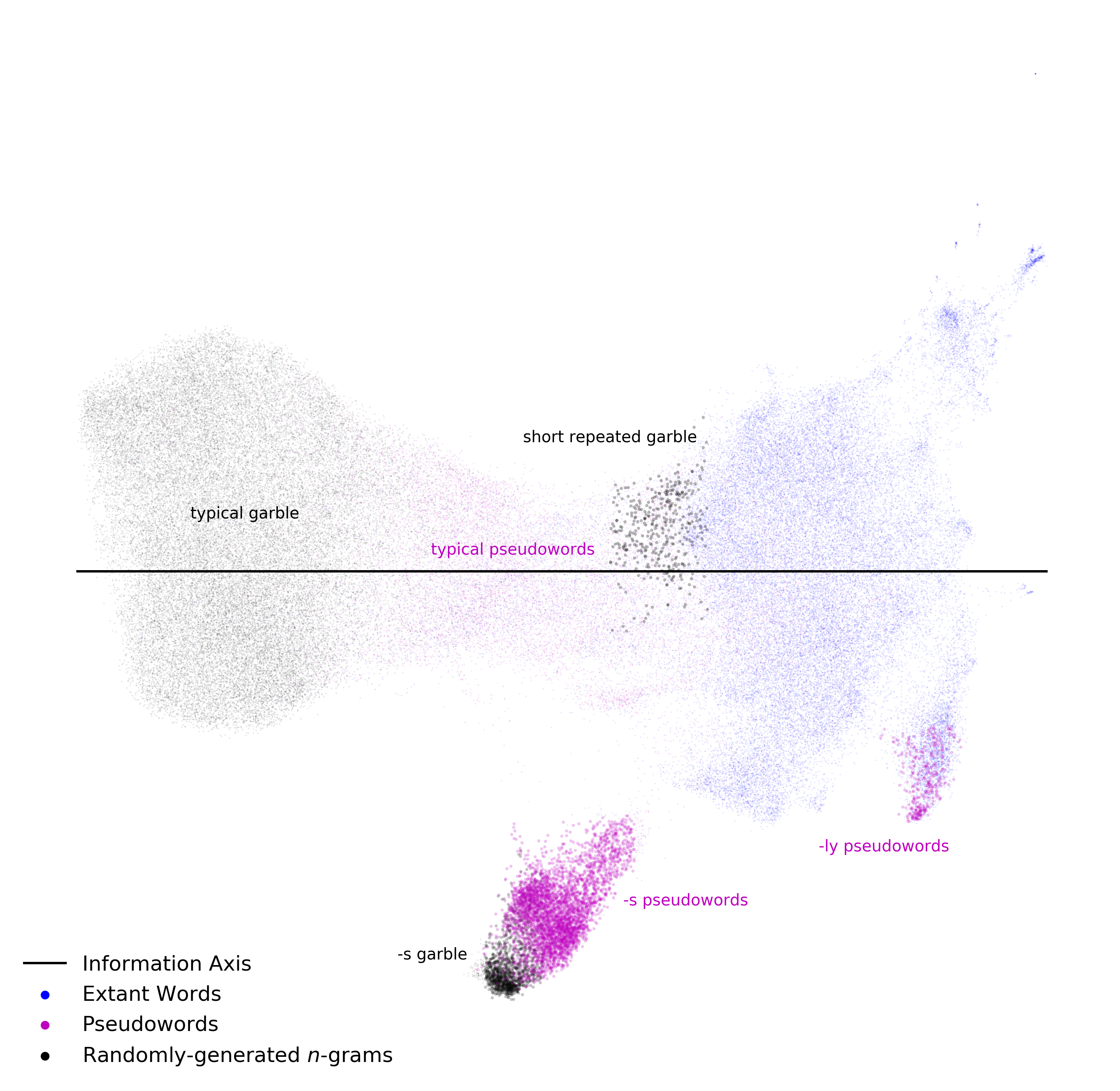}
    \caption{UMAP projection of CharacterBERT embeddings for extant words (blue), pseudowords (magenta), and randomly generated character $n$-grams (black). The solid black line shows the information axis that we define in this work. We discuss the highlighted clusters of pseudoword and garble $n$-grams in Appendix \ref{appendixb}.}
    \label{fig:fig7}
\end{figure*}

Here, we qualitatively explore the main features of pseudoword and randomly generated $n$-grams' structure in our UMAP projection, deferring a more detailed exposition to future work. Figure \ref{fig:fig7} highlights several distinct groups of randomly generated (black) and pseudoword (magenta) $n$-grams that we describe in detail below.

Beginning with randomly generated $n$-grams, we first note that there is a significant correlation between their string length and information axis score, such that randomly generated $n$-grams with low information axis scores tend to contain more characters, and vice versa. Indeed, the high-information tail of the garble distribution shown in Figure \ref{fig:fig2} has a power-law exponent that is quantitatively consistent with the low-length tail of the underlying string length distribution.\footnote{We remind the reader that our randomly generated string length distribution is matched to that of our extant sample.} Figure \ref{fig:fig7} highlights two notable exceptions to this rule: a cluster of randomly generated $n$-grams with strings that tend to be short and often contain repeated characters, and a garble cluster in which strings tend to end in ``s.'' We refer to the remaining randomly generated $n$-grams as ``typical garble.'' To illustrate, we provide ten examples of $n$-grams in each category:
\begin{itemize}
    \item Typical garble: kiwbckodaffzhjxkvpfh, ijhtsfjsu, ojcfere, fsgnwy, qiqa, nevm, uzp, tgj, bv, w;
    \item Short repeated garble: cureuul, fbxoon, gallm, alln, ffod, ido, obb, tek, aa, hq;
    \item -s garble: dddgvasbbzaoeuius, wdycrynylhyos, nkeccmosls, ilvtubdts, eoubazos, ptfjqs, hslxls, xwkss, gehs, jgs.
\end{itemize}

A particularly interesting feature of short repeated garble is that it encodes considerably more information along our axis than the pseudowords in our sample. This is striking because the pseudowords were generated using an algorithm designed to generate morphologically plausible words, whereas the garble is generated purely randomly at the character level. In this way, our garble embeddings provide novel insights into the string patterns that CharacterBERT identifies as information rich and predictive of word plausibility (in terms of proximity to extant words in embedding space); specifically, it reveals that CharacterBERT identifies repeated characters in the same string as information rich, even though these repeated character sequences often lack morphological hallmarks of extant words.

These sequences of repeated sounds share similarities with early-childhood vocalizations (“babbling”), as well as stylistic features of child-oriented speech in the context of early word learning, for example, words such as \textit{mama} and \textit{dada}. The role of simple character repetitions in child development is often studied from the phonetic standpoint as a mechanism for a child to become proficient in the diversity of sounds appearing in a language \cite{DavisMacNeilage1995}. However, the results from CharacterBERT suggest that repetitive sequences are especially rich in their information at the character-level, which may confer additional syntactic or lexical benefits as children learn to differentiate random sounds from linguistically meaningful units. We leave further investigation of this to future work.

Pseudoword embeddings also display a clear cluster in which strings tend to end in ``s.'' In addition, there is a distinct pseudoword group near extant adverbs (see Figure \ref{fig:fig3}) in which strings tend to end in ``ly.'' We refer to the remaining pseudoword $n$-grams as ``typical pseudowords.'' To illustrate, we provide ten examples of $n$-grams in each category:
\begin{itemize}
    \item Typical pseudowords: hypnostementer, eatmendownwald, eninardister, unalgion, conquing, ambooked, runton, ditity, etbarn;
    \item -s pseudowords: sacrembelcones, irstuphorries, unnessnells, herepairds, finihips, littoes, warposs, quards, prects, gicass;
    \item -ly pseudowords: queepecturusly, unbornortardly, remechlocally, expotputtly, musteetly, confully, popubly, ectoily, artfaly, mously.
\end{itemize}

\end{document}